\documentclass[letterpaper]{article} 
\usepackage[]{aaai23}  
\usepackage{times}  
\usepackage{helvet}  
\usepackage{courier}  
\usepackage[hyphens]{url}  
\usepackage{graphicx} 
\usepackage{natbib}  
\usepackage{caption} 
\usepackage{algorithm}
\usepackage{algorithmic}

\usepackage{subcaption}
\usepackage{newfloat}
\usepackage{listings}
\DeclareCaptionStyle{ruled}{labelfont=normalfont,labelsep=colon,strut=off} 
\floatstyle{ruled}
\newfloat{listing}{tb}{lst}{}
\floatname{listing}{Listing}
\usepackage{amsfonts,amsmath,algorithm}
\usepackage{amssymb}
\usepackage{bm}
\usepackage{mathabx}

\usepackage{wrapfig}

\usepackage{graphicx}
\usepackage{caption}
\usepackage{float}

\usepackage{xspace}
\usepackage{multirow}
\usepackage{mathtools}
\usepackage{physics}
\usepackage{cancel}

\RequirePackage[textsize=scriptsize]{todonotes}

\usepackage{thmtools}
\usepackage{thm-restate}

\newcommand\Mycomb[2][^n]{\prescript{#1\mkern-0.5mu}{}C_{#2}}

\newcommand{\indep}{\bot\!\!\!\!\bot}

\usepackage{subcaption}
\usepackage{caption}

\NewDocumentCommand{\tens}{e{_^}}{%
  \mathbin{\mathop{\otimes}\displaylimits
    \IfValueT{#1}{_{#1}}
    \IfValueT{#2}{^{#2}}
  }%
}

\makeatletter
\newcommand*{\centernot}{%
  \mathpalette\@centernot
}
\def\@centernot#1#2{%
  \mathrel{%
    \rlap{%
      \settowidth\dimen@{$\m@th#1{#2}$}%
      \kern.5\dimen@
      \settowidth\dimen@{$\m@th#1=$}%
      \kern-.5\dimen@
      $\m@th#1\not$%
    }%
    {#2}%
  }%
}
\makeatother

\makeatletter
\DeclareRobustCommand\notindep{\mathrel{\m@th\mathpalette\c@ncel\indep}}
\makeatother

\newcommand{\red}[1]{{\color{red} #1}}

\title{Progressive Fusion for Multimodal Integration}
\author{
    Shiv Shankar,
    Laure Thompson,
    Madalina Fiterau
}
\affiliations{
    University of Massachusetts\\
    \{sshankar,laurejt,mfiterau\}@cs.umass.edu

}

\begin{document}

\maketitle

\begin{abstract}

Integration of multimodal information from various sources has been shown to boost the performance of machine learning models and thus has received increased attention in recent years. Often such models use deep modality-specific networks to obtain unimodal features which are combined to obtain ``late-fusion" representations. However, these designs run the risk of information loss in the respective unimodal pipelines. On the other hand, ``early-fusion" methodologies, which combine features early, suffer from the problems associated with feature heterogeneity and high sample complexity.
In this work, we present an iterative representation refinement approach, called Progressive Fusion, a model-agnostic technique which makes late stage fused representations available to early layers through backward connections, improving the expressiveness of the  representations. Progressive Fusion avoid the information loss which occurs when late fusion is used, while retaining the advantages of late fusion designs. We test Progressive Fusion on tasks including affective sentiment detection, multimedia analysis, and time series fusion with different models, demonstrating its versatility. We show that our approach consistently improves performance, for instance attaining a 5\% reduction in MSE and 40\% improvement in robustness on multimodal time series prediction.
 
\end{abstract}

\section{Introduction}
%

Traditionally, research in machine learning has focused on different sensory modalities in isolation, but it is well recognized that human perception relies on the integration of information from multiple sensory modalities.
Multimodal fusion research aims to fill this gap by integrating different unimodal representations into a unified common representation \citep{turchet2018internet,baltruvsaitis2018multimodal}. 
Typically, fusion techniques fall into two categories, \emph{early fusion} and \emph{late fusion}, depending on where the information from each modality is integrated in the feature pipeline  \citep{varshney1997distributed, ramachandram2017deep}. While theoretically early fusion models tend to be more expressive, in practice they are more commonly used for homogeneous or similar modalities \citep{ramachandram2017deep}. On the other hand late fusion models are more effective in combining diverse modalities. This has generally been attributed to the challenges like feature shifts, cross-modal distributional changes, difference in dimensionality, etc. when dealing with heterogeneities across diverse modalities such as text and image \citep{mogadala2021trends, yan2021deep}. 

In this work, we aim to bridge this divide by using backward connections which connect the late fused representation ( \`{a} la late fusion) to unimodal feature generators, thus providing cross-modal information to the early layers ( \`{a} la early fusion).
This creates a model that learns to progressively refine the fused multimodal representations. 

We show that our proposed technique called progressive-fusion (Pro-Fusion) results in improvements of different multimodal fusion architectures including recent \emph{state of the art models} such as MAGXLNET \citep{rahman2020integrating}, MIM \citep{han2021improving} and MFAS \citep{perez2019mfas}. Our experiments show that training with the Pro-Fusion design results in more accurate and robust models compared to the initial state of the art architectures.


\textbf{Contributions:}
(1) We propose a framework to bridge the gap between early and late fusion via backward connections.
(2) We apply this model-agnostic approach to a broad range of state of the art models for a diverse set of tasks.
(3) We show, through rigorous experiments, that models trained with Pro-Fusion are not just consistently more accurate, but also considerably more robust than the corresponding standard baseline models. We show up to 2\% improvement in accuracy over state of the art sentiment prediction models and up to 5\% reduction in MSE and 40\% improvement in robustness on a challenging multimodal timeseries prediction task.


\section{Preliminaries and Related Work}

\subsection{Multimodal Fusion}
\label{sec:prelim}
Multimodal learning is a specific type of supervised learning problem with different types of input modalities. We are provided with a dataset of $N$ observations $\mathcal{D} = { (X^j,Y^j)_{j=1}^{N}} $, where all $X^j$ come from a space $\mathcal{X}$ and $Y^j$ from $\mathcal{Y}$, and a loss function $L : \mathcal{Y} \times \mathcal{Y} \rightarrow \mathbb{R}$ which is the task loss. Our goal is to learn a parametric function $\mathcal{F}: \mathcal{X} \rightarrow \mathcal{Y}$ such that the total loss  $\mathcal{L} =  \sum_j L(\mathcal{F}(X^j),Y^j)$ is minimized. In multimodal fusion the space of inputs $\mathcal{X}$ naturally decomposes into $K$ different modalities $\mathcal{X} = \prod_{i=1}^K \mathcal{X}_i$. Correspondingly any observation $X^j$ also decomposes into modality specific components $X^j_i$ i.e. $X^j = (X^j_1,X^j_2, \dots X^j_K) $.

A natural way to learn such a function with a multimodal input is to have an \emph{embedding component}  which fuses information into a high dimensional vector in $\mathbb{R}^d$, where $d$ is the size of the embedding, and a \emph{predictive component} $P$ which maps the embedding vector from $\mathbb{R}^d$ to $\mathcal{Y}$. Furthermore, since different modalities are often of distinct nature and cannot be processed by similar networks (e.g. text and image), the embedding generator is decomposed into (a) unimodal feature generators $G_i : \mathcal{X}_i \rightarrow \mathbb{R}^{d_i}$ which are specifically designed for each individual modality $\mathcal{X}_i$ and (b) a fusion component $F : \prod_i \mathbb{R}^{d_i} \rightarrow \mathbb{R}^d$ which fuses information from each individual unimodal vector. $F$ is provided with unimodal representations of the input $X^j$ obtained through embedding networks $G_i$. The unimodal feature generators $G_i$ can have different kinds of layers including 2D convolution, 3D convolution and fully connected layers. $F$ is the layer where the embeddings obtained from different modalities are fused. $F$ is called the \emph{fusion} or \emph{shared representation} layer. $F$ has to capture both unimodal dependencies (i.e. relations between features that span only one modality) and multimodal dependencies (i.e. relationships between features across multiple modalities).


\subsection{Prior Approaches to Fusion}

\begin{figure*}[th!]
    \centering
    \begin{subfigure}[b]{0.31\textwidth}
    \centering
    \includegraphics[width=\textwidth]{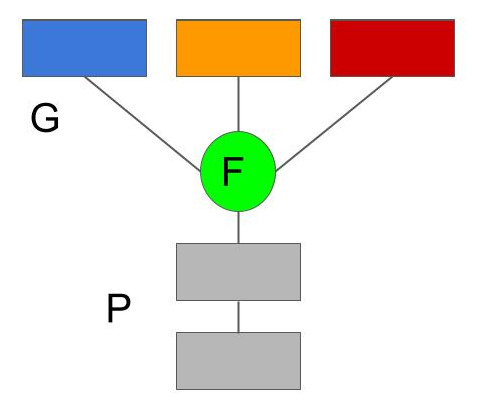}
    \caption{Early Fusion \label{fig:base_fusion:early}}
    \end{subfigure}
    ~
    \begin{subfigure}[b]{0.31\textwidth}
    \centering
    \includegraphics[width=\textwidth]{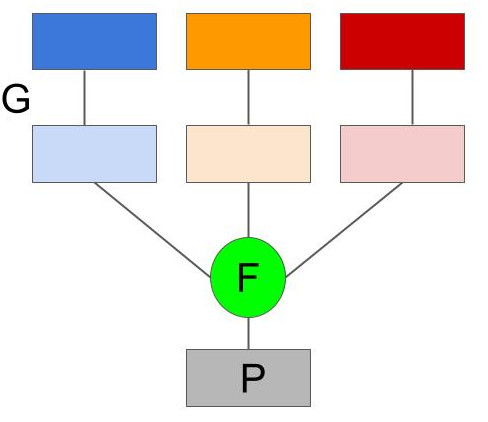}
    \caption{Late Fusion \label{fig:base_fusion:late}}
    \end{subfigure}
    ~
    \begin{subfigure}[b]{0.31\textwidth}
    \centering
    \includegraphics[width=\textwidth]{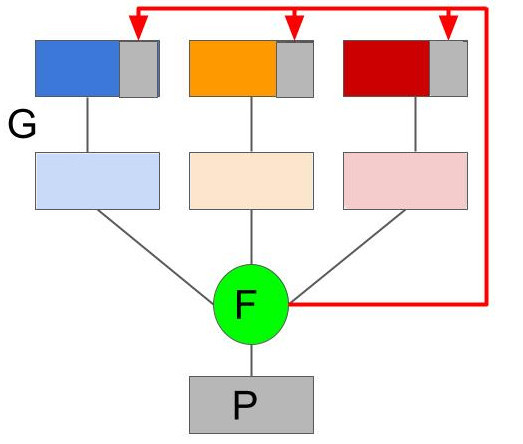}
    \caption{Pro-Fusion \label{fig:base_fusion:rec}}
    \end{subfigure}
    \caption{Representative Multimodal Fusion Architectures of a) Early fusion , b) Late fusion and c) Pro-Fusion. We have also indicated the components mentioned in Section \ref{sec:prelim} viz. the unimodal feature generators $G$, fusion layer $F$ and predictive network $P$ in the figures. Generally models with high capacity $P/G$ are considered early/late fusion respectively. The key difference between a late fusion architecture and pro-fusion architecture are the skip-back connections, indicated in red.}
    \label{fig:base_fusion}
\end{figure*}

\textbf{Early Fusion and Late Fusion} Early and Late Fusion is a broad categorization of fusion models in terms of how the fusion operation (layer where features from different modalities are combined) splits the complexity of the model \citep{ayache6, ramachandram2017deep}. Early fusion ( Figure \ref{fig:base_fusion:early}) combines the modalities `early' and devotes more model complexity to post-fusion operations. In contrast to early fusion, late fusion design ( Figure \ref{fig:base_fusion:late} ), combines the features from different modalities at a `late' stage. Early fusion models have the ability to model highly complex dependencies between different modalities, but face problems when dealing with heterogenous sources such as text and images. Late fusion allows an easy way to aggregate information from diverse modalities and run the risk of missing cross-modal interactions in the mixed feature space.

\textbf{Architecture Changes} Many recent works including that of \citet{vielzeuf2018centralnet}, \citet{sankaran2021multimodal}, \citet{perez2019mfas}, \citet{hazarika2020misa} design new deep architectures. \citet{vielzeuf2018centralnet} proposed a CentralNet design based on aggregative multi-task learning. \citet{sankaran2021multimodal} design a Refiner Fusion Network (Refnet) trained via cyclic losses. \citet{perez2019mfas} used neural architecture search to find a good architecture for convolutional networks. \citet{hsu2018disentangling} and \citet{khattar2019mvae} use multimodal autoencoders to learn better representations. \citet{tsai2019multimodallearning} improved upon the factor model based approach of \citet{hsu2018disentangling}. \citet{nagrani2021attention}  modify the multimodal transformer \citep{tsai2019multimodal} to incorporate bottlenecks.

Our proposed method, though technically an architecture change, is a single change that \emph{treats the existing model as given}. It is closer in spirit to a black-box change, compared to the aforementioned methods. Hence it is \emph{complementary} to this line of work. We experiment with many of the aforementioned models to show how our proposal consistently improves performance.

\textbf{Fusion Techniques}
Other than basic fusion layers such as pooling and concatenation, other common layers used include aggregation \citep{khan2012color}, tensor factorisation \citep{liu2018efficient, zadeh2017tensor}, attention \citep{tsai2019multimodal} and memory modules \citep{zadeh2018memory}. \citet{rahman2020integrating} design a model using pre-trained transformer to achieve state of the art results on the multimodal sentiment benchmarks. These works propose specific fusion techniques, they design specific forms of the $F$ function (see Figure \ref{fig:base_fusion}). Our proposed technique is \emph{agnostic to the choice of the fusion function $F$} and is \emph{orthogonal} to these ideas. 

\textbf{Model Agnostic Methods}
Model independent methods to improve fusion by using train-time objectives based on cortical synergy \citep{shankar2021neural}, mutual information \citep{colombo2021improving, bramon2011multimodal}, or contrastive estimation \citep{Liu_2021_ICCV} ave been widely explored.
Recently \citet{wang2020makes} proposed a reweighing based approach (GB) to tackling the problem of jointly learning models of differing capacity. Our proposal is distinct from these methods in that it adds backprojective connections. These model-agnostic proposals are generally orthogonal to our approach, and potentially can be combined \emph{to achieve further improvements}. For example, in our experiments we will show that our method can increase performance on the model-agnostic GB based approaches as well.



\section{Progressive Fusion (Pro-Fusion)}


\subsection{Motivating Example}
\label{sec:motiv}

\begin{figure}[!htb]
    \centering
    \begin{subfigure}[b]{0.45\textwidth}
    \centering
    \includegraphics[width=\textwidth]{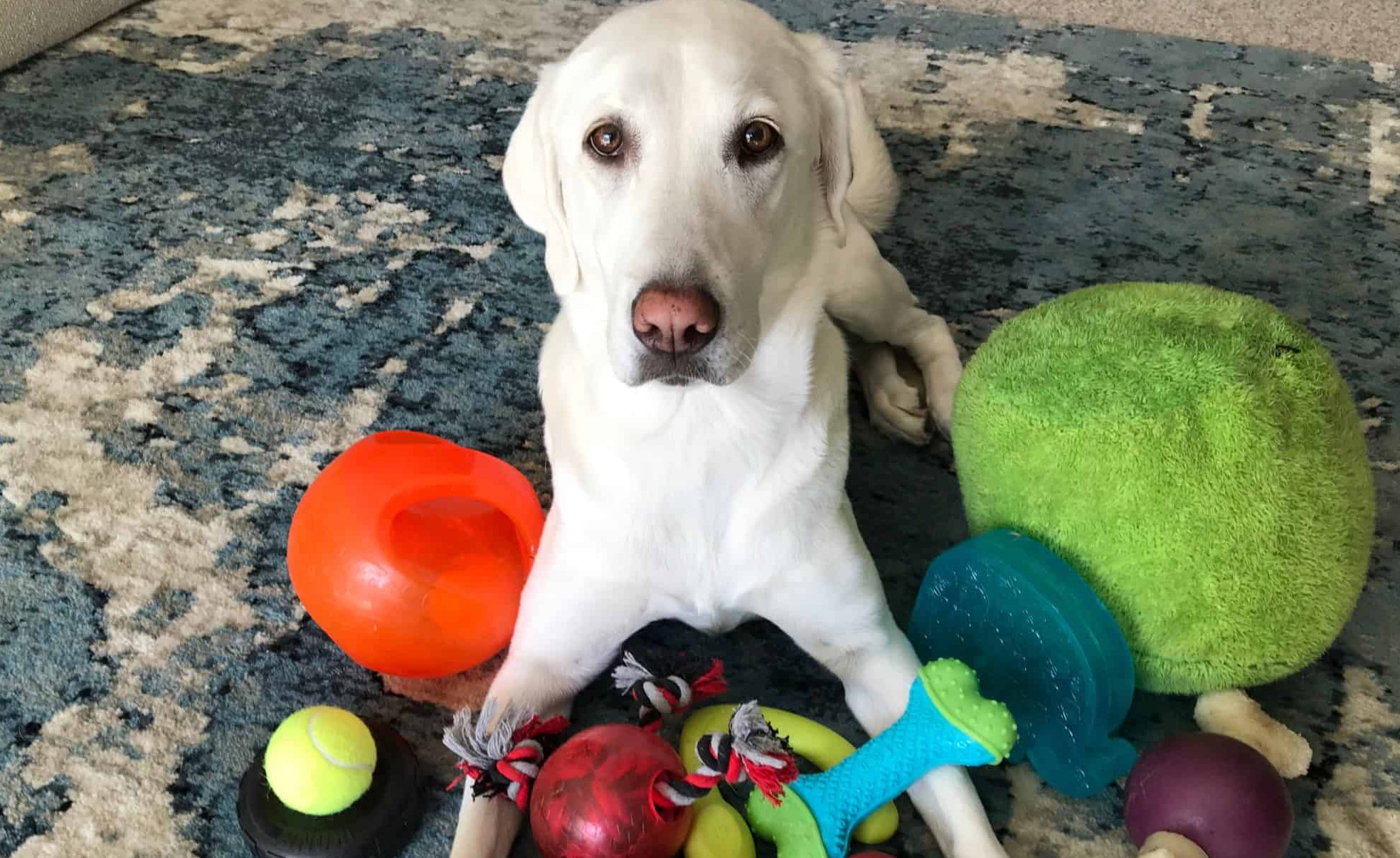}
    \caption{Example image for motivating case. The target corresponds to the location in the image of the object described in the audio modality (dog, ball, bone etc). Source: \url{https://kylekittleson.com/}}.
    \label{fig:motiv} 
    \end{subfigure}
    ~
    \begin{subfigure}[b]{0.44\textwidth}
    \includegraphics[width=0.8\textwidth]{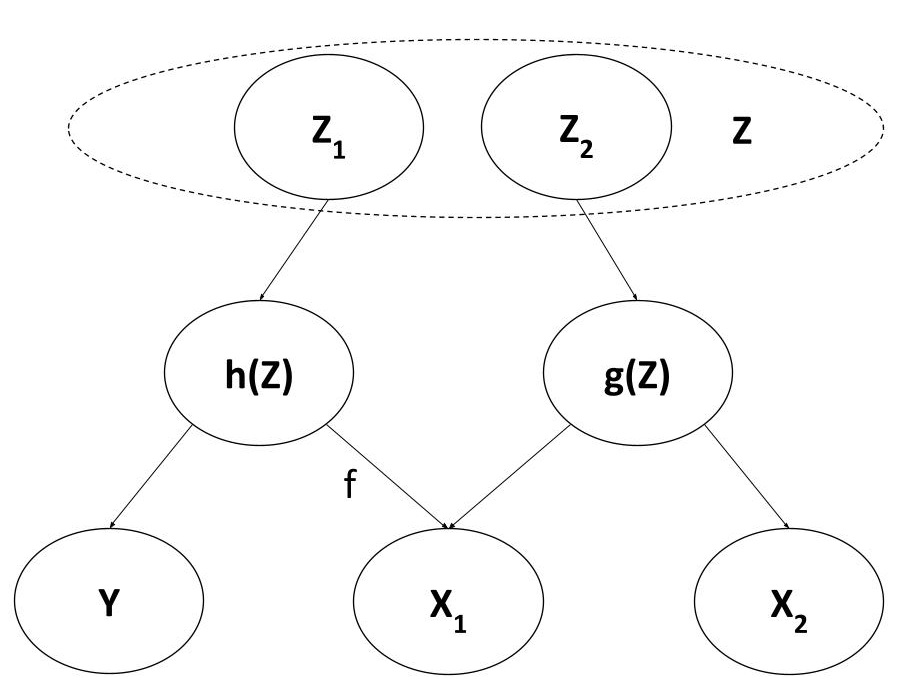}
    
        \caption{ Generative Model for the case discussed in the `Motivating Example' section. $Z$ is the latent vector that determines the outcome $Y$ via $h(Z)$. $g(Z)$ is a function of $Z$ independent of $Y$. $X_1$ is a combination of $h(Z) \text{ and } g(Z)$.
    \label{fig:M1}}
    \end{subfigure}
\end{figure}

Consider the task of determining the location of an entity from video and text. For instance, suppose the system has to detect the coordinates, in a given image, of an object specified through a textual command. For the image provided in Figure~\ref{fig:motiv}, the text might be `find the tennis ball' or `find the blue bone'. The task is not solvable using a single modality, as the image only contains the objects and their location, whereas the text only mentions the object of interest.

Now consider what might happen with a late-fusion scheme. A significant part of the representation capacity of the image features might be devoted to capturing the dog, the bone, the carpet etc. And hence determining the coordinates of the small tennis ball will be more challenging unless the image feature generator has access to the textual information. More generally, if the unimodal feature generators are bottlenecked or not powerful enough; the required information to predict the output might be lost or compressed too much to be recovered correctly. With early fusion, the image feature generator knows which object to focus on and can be directed towards the relevant information, specifically the location and other information about the tennis ball.

Figure \ref{fig:M1} represents an abstract graphical model for this situation. $X_1$ represents the entire input image, while $Z$ represents an abstract state of the environment (with objects and coordinates). The output $Y$ (e.g. coordinate target) is determined by the feature function $h$ so $Y \leftarrow h(z)$ ( i.e. $h(Z)$ contains sufficient statistics about  the location of the object). The information about these features are present in $X_1$ (obtained by applying unknown function $f$ to $h(Z)$); however $X_1$ has nuisance variables (e.g. other objects) or a corrupted version of $h(z)$. $g(Z)$ represents descriptions like colour, shape etc of the target object. The location $h(Z)$ and visual characters $g(Z)$ put together forms (part of) the image. In this case $Y$ may not be identifiable purely via $X_1$ . For the image example, this is because $X_1$ has not just the target but other objects which means the without characterizing the desired target, a specific location cannot be specified. But in the presence of input $X_2$  (in this case text) the target is identifiable \textit{even} if $X_2$ by itself is not informative about $Y$. \footnote{For an example in terms of equations refer to the Appendix \ref{apx:synth}}

In such a scenario, with late fusion based approach if the encoder $G_1$ (unimodal feature generator for mode $\mathcal{X}_1$) is not sufficiently powerful, the overall networks may not be able to learn the perfect function $f$ even in the presence of modality $\mathcal{X}_2$. This can happen during late fusion when the network $F_1$ has already pooled together $h$ and $g$ in a non-invertible manner. On the other hand if the features from $X_2$ were made available to the feature encoder for $X_1$, it can learn to ignore or mask nuisance variation/corruption. This however requires the model to perform some form of early fusion; but if the underlying modalities $\mathcal{X}_2$ and $\mathcal{X}_1$ are very different, the corresponding integration is challenging. 

More generally, fusion at higher level features run into a “fuse it or lose it” situation where relevant information -- especially conditionally relevant information -- that is not fused by the fusion layer is at risk of being lost.  From the motivating example, only in presence of $X_2$ (speech command) could we process $X_1$ (image) to get $h(Z)$ (location). If location information in $X_1$ is corrupted before the fusion layer, the problem becomes intractable. This is happening because the unimodal feature generation is unaware of features coming from other modalities. Early fusion does not face this problem, but cannot handle heterogenous modalities well, and requires a huge number of parameters. This leads us to our basic problem: of design a generic approach that combines the advantages of late and early fusion. To this end, we propose a model agnostic scheme based on providing late-stage multi-modal fusion features to the early stages of unimodal feature generators. 

\subsection{Pro-Fusion}
\label{sec:main}
We build a scheme based on backprojective connections which can be applied to any given base architecture. 
Our scheme considers any given base design as a single step of an iterative process. The base design is augmented to take an additional context vector as input, which serves to provide information from `late' fused representations.
At each iteration, the current output representations of the base model are provided via the context vector as an additional input for the next step.  More formally, given a base model $\mathcal{F}$ with input $x = (x_i,x_2,..x_k)$, we want to create an augmented model $\hat{\mathcal{F}} : \mathcal{X} \times \mathbb{R}^{d} \rightarrow \mathcal{Y}$ with additional input $c \in \mathbb{R}^{d}$ such that $ c = 0 \implies \hat{\mathcal{F}}(x,c) = \mathcal{F}(x)$. Recall that the function $\mathcal{F}$ mentioned in Section \ref{sec:prelim} is given by $
\mathcal{F}(x) = P(F(G_1(x_1),G_2(x_2),..G_K(x_K)))
$. 

\begin{figure*}[t]
    \centering
    \includegraphics[height=0.15\textheight, width=0.9\textwidth]{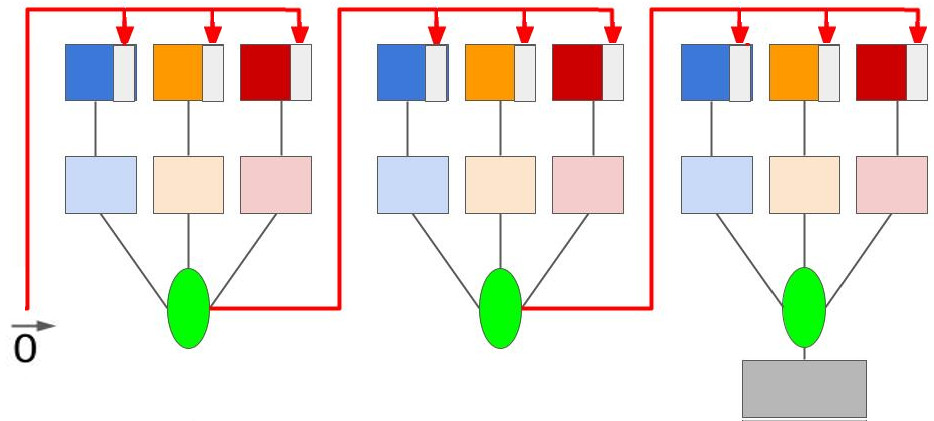}
    \caption{ Example Pro-Fusion design unrolled for 3 steps with the Late Fusion Model (Figure \ref{fig:base_fusion:late}).}
    \label{fig:rec_fusion}
\end{figure*}

We create the desired network $\hat{\mathcal{F}}$ by providing $c$ to the unimodal feature generators $G_j$. We use the output of the fusion layer $F$ and project it back into the network as $c_t$ via the matrix/function $W_i$. This creates an iterative network which we run for $R$ steps. The final vector $c_R$ after $R$ steps serves as the output of fusion which is then provided to the predictor model $P$.  In Figure \ref{fig:rec_fusion} we visually depict how the model is unrolled.

The ability to send information backward in the network addresses the problem raised earlier in Section~\ref{sec:motiv} 3. The encoder $G_1$ for $X_1$ can now gain a handle on $g(Z)$ via the fused output $c_1$. Specifically if the model can compute $g(z)$ from $W(c_1)$, then in the second iteration step, one can recover from $X_1$ the real value of $h(Z)$, which then directly determines the target $Y$. On the other hand if $X_2$ is not useful or if $G_1$ cannot process the fused vector efficiently, then $W(.)$ can be zeroed out and the overall model is no worse than the baseline model.
We also present in the Appendix \ref{apx:theory}, some mathematical analysis as to the representation power of our approach.

The importance of multimodal backward connections can also be interpreted from the perspective of the graphical model in Figure \ref{fig:synth_cont}. A standard message passing routine \citep{koller2009probabilistic} on the aforementioned graph, will have the message from $X_2$ effecting the belief of target $Y$ via two paths: a) one along $X_2, g(Z), Z, h(Z)$ and the other along $X_2, g(Z),X_1, h(Z)$. Notice that along this second path, message from the modality $X_2$ is received at $X_1$ before further processing. This path makes features from modality $X_2$ available to the feature generator of $X_1$, which is exactly what the backprojection layer also accomplishes. One caveat here is that unlike this example, in general we do not know which way to route the messages (as the dependence graph maybe unknown). As such in our proposal we treat all modalities symmetrically and re-cycle information through all of them.

An astute reader might notice similarities with deep unfolding networks \citep{balatsoukas2019deep,hershey2014deep}. However, these are not designed for multimodal data, nor adapted to it, to the best of our knowledge. In contrast, ProFusion was specifically designed to solve a problem in multimodal data fusion: the “fuse it or lose it” situation. Deep unfolding/iterative models that do not cycle cross-modal information still suffer from the “fuse it or lose it” problem. This can be seen in our experiments (Section 4.5) where we show that ProFusion provides additional improvement over deep unrolling style iterative models. Secondly, unrolling is just one method to train the backward connections. We refer the readers to Appendix 1, for an expanded discussion on this.





\section{Experiments}
\label{sec:expts}
In this section, we empirically demonstrate that Pro-Fusion improves performance of multimodal deep learning SOTA architectures on a variety of tasks. First we verify our intuition for the advantage of backward connections in a synthetic experiment.  Next we experiment with datasets in sentiment prediction \citep{zadeh2018multi}, multimedia classification \citep{vielzeuf2018centralnet} and financial timeseries prediction \citep{sardelich2018multimodal}. 
We also explore how our approach affects robustness for noisy time series data. Finally we evaluate the impact of varying the number of unrolling steps and analyze how the model performance as well as unimodal representations evolve. 
For all the datasets we use SOTA and near-SOTA models, while keeping a diversity of fusion techniques and network designs. For each dataset and architecture combination, we either use established hyperparameters and/or choose the best hyperparameter from our own experiments. Next for the same architecture, we \emph{add backward connections} from the fusion layer output and train with the exact same hyperparameters. We \emph{do not perform any hyperparameter tuning or search for our modified models}, so the reported results are a lower bound on what Pro-Fusion can achieve. We opt for this process to 
isolate the effects of adding backward connections from those of tuning hyperparameters.

\begin{figure}[!htb]
\centering
\includegraphics[width=0.4\textwidth,keepaspectratio=true]{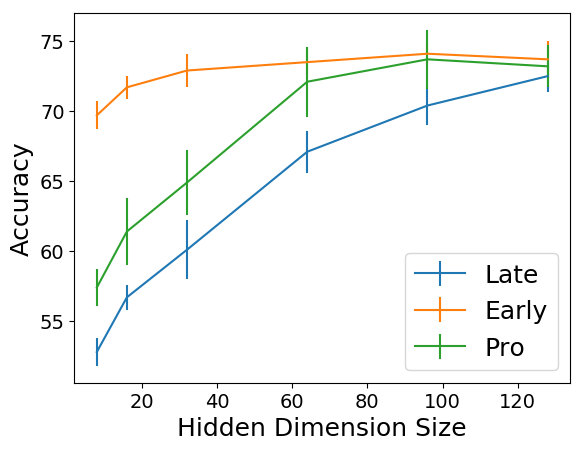}
\caption{Accuracy of late, early and pro-fusion models over varying levels of inner dimension. Each point corresponds to the performance of the corresponding model when the hidden dimension is set to the values of $d$ on the x axis.
\label{fig:synth_new_dim}}
\end{figure}
\begin{table}
\centering
\begin{tabular}{|l|l|l|}
\hline
 & \multicolumn{2}{c|}{Accuracy $\uparrow$} \\
\hline
Model  &  Base &  Ours \\
\hline
\hline
LF     & 71.4    &   71.6        \\
LFN    & 71.1     &   \textbf{71.8}        \\
MFM    & 71.4     &  \textbf{72.2}         \\
GB   & 68.9     &   69.3               \\
Refnet & 70.6    &  \textbf{71.2}          \\
MFAS & 72.1     &  \textbf{72.5}           \\
MBT & 70.3      &  70.3 \\
\hline
\end{tabular}
\caption{Results on digit classification task with AVMNIST for various fusion architectures. The performance metric is Accuracy, and was measured on five trials. Our method \emph{outperforms the baseline in almost all instances}. Scores above 1 standard deviation of the base models, indicating significance, have been highlighted. \label{tab:avmnist}}
\end{table}

\subsection{Synthetic Dataset}
To verify the intuition described in the `Motivating Example' Section, we first create a synthetic experiment. For this purpose we encode a smooth random function in modality $\mathcal{X}_1 \subset \mathbb{R}^D$. Specifically the $d^{\text{th}}$ component of $X_1$ has the value of the function at the point $d/D$. Next, in modality $\mathcal{X}_2$, we provide position embeddings of a randomly chosen lattice point $ l \in \{0,1/D,2/D,...1\}$. The output label $Y$ is the first non-zero digit of $x_l$. This is conceptually a simple task as one can infer the component from the modality $X_2$  and simply read on the corresponding component from $X_1$. However if the model is late fusion, where the input modalities might go through a lower dimensional representation; the specific values of each component in $X_1$ is lost, and the model cannot correctly predict the label. Note that in this case, each instance of $X_1$ contains a different function; because a fixed function might be directly learned by the network.

In Figure \ref{fig:synth_new_dim}, we plot the accuracy of a 2 layer MLP trained on this task with different sizes for the hidden layer. The argument from Section 3.1 suggests that early fusion is more effective than late fusion when the hidden layers are smaller. Furthermore it also suggests that the effect of progressive fusion is larger when the feature layers input to late fusion is smaller. This is confirmed by the experiments, where the gap between the pro-fusion model and late fusion model reduces as the size of the hidden representation increases. Finally for a large enough hidden representation, the performance of late fusion matches that of early fusion. Additional analysis using synthetic data is shown in the Appendix. 

\subsection{Multimedia Classification}

\textbf{Datasets.}
We first evaluate our proposed design changes on AV-MNIST \citep{vielzeuf2018centralnet}, a popular benchmark dataset used for multimodal fusion \citep{perez2019mfas,joze2020mmtm}. It is an audio-visual dataset for a digit classification task. The data is prepared by pairing human utterances of digits obtained from FSDD dataset \footnote{\url{https://www.tensorflow.org/datasets/catalog/spoken_digit}} with images of written digits from MNIST. This dataset has 55K training, 5K validation, and 10K testing examples.
To prepare the dataset we use the processing stack of \citet{cassell20mfas}. The preprocessing involves adding corruption to both modalities, so that no single modality is sufficient \citep{vielzeuf2018centralnet}. 


\textbf{Models.}
\textbf{LF} is the baseline late fusion architecture used in \citet{vielzeuf2018centralnet}. \textbf{MFAS} is the architecture search based model used by \citet{perez2019mfas}. It is the current SOTA on AV-MNIST. The exact architecture is presented in the Appendix \ref{apx:mfas_avmnist}. We use the model obtained by search and add the backward connections. \textbf{LFN} is the low rank tensor fusion approach \citep{zadeh2017tensor} adapted to this dataset, while  \textbf{MFM} refers to the factorization method of \citet{tsai2019multimodallearning} for learning multimodal representation. \textbf{GB} and \textbf{Refnet} are the gradient blending and refiner network based approaches of \citet{wang2020makes} and \citet{sankaran2021multimodal} respectively. \textbf{MBT} is the multimodal transformer model of \citet{nagrani2021attention}. Similar to existing work \citep{vielzeuf2018centralnet, liang2021multibench} our experiments used the aforementioned fusion designs with unimodal LeNet style feature generators, except for MBT which uses transformers.

Our results are presented in Table \ref{tab:avmnist}.
Amongst all the methods we evaluated, Pro-MFAS was the best model and beats its standard counterpart by 0.4 accuracy points. We also observe similar improvements in using Pro-Fusion with the MFM design. In fact the Pro-fusion MFM model was competitive with the current state of the art MFAS model. Meanwhile, the gradient blending (GB) based fusion approach seems to not generalize on this dataset and  performs worse than even the baseline late fusion method.

\subsection{Sentiment Prediction}
\textbf{Datasets.}
We empirically evaluate our methods on two datasets CMU-MOSI \citep{wollmer2013youtube} and CMU-MOSEI \citep{zadeh2018multi}. CMU-MOSI is sentiment prediction tasks on a set of short youtube video clips. CMU-MOSEI is a similar dataset consisting of around 23k review videos taken from YouTube. Both of these are used generally for multimodal sentiment analysis experiments. Audio, video, and language modalities are available in each dataset.

\textbf{Models.} \textbf{FLSTM} is the early fusion type baseline LSTM architecture used by \citet{zadeh2017tensor}, while \textbf{LFN} is the low rank tensor representation of model of \citet{zadeh2017tensor}. multimodal features.  \citep{hazarika2020misa}. \textbf{MAGBERT} and \textbf{MAGXLNET} \citep{rahman2020integrating} are BERT \citep{devlin2018bert} based state of the art models on these datasets. These architectures uses a gating mechanism \citep{wang2019words} to augment a pretrained transformer. \textbf{MIM} \citep{han2021improving} is a recent near state of the art architecture. It combines BERT based text embeddings with modality specific LSTMs.

\begin{table}
\centering
\small
\begin{tabular}{|l|c|c|c|c|}
\hline
         & \multicolumn{2}{c|}{$Acc_7 \uparrow$} & \multicolumn{2}{c|}{$Acc_2 \uparrow$}  \\ 
\hline
 Model        & Base & Ours                           & Base & Ours                            \\
 \hline
 \hline
FLSTM    & 31.2 & \textbf{31.8}                          & 75.9 & \textbf{76.8}                  \\
LFN      & 31.2 & \textbf{32.1}                          & 76.6 & \textbf{77.2}                  \\
MAGBERT  & 40.2 & \textbf{40.8}                          & 83.7 & \textbf{84.1}                  \\
MAGXLNET & 43.1 & \textbf{43.5}                          & 85.2 & 85.5                           \\
MIM      & 45.5 & \textbf{46.3}                          & 81.7 & \textbf{83.4}                  \\
\hline                         
\end{tabular}
\caption{Results on sentiment analysis on CMU-MOSI. $Acc_{7}$ and $Acc_{2}$ denote accuracy on 7 and 2 classes respectively. Boldface denotes statistical significance. \label{tab:mosi}}
\end{table}

\begin{table}
\centering
\small
\begin{tabular}{|l|c|c|c|c|}
\hline
         & \multicolumn{2}{c|}{$Acc_7 \uparrow$} & \multicolumn{2}{c|}{$Acc_2 \uparrow$}  \\ 
\hline
 Model        & Base & Ours                           & Base & Ours                            \\
 \hline
 \hline
FLSTM    & 44.1 & \textbf{44.8}                          & 75.1 & \textbf{75.8}                  \\
LFN      & 44.9 & \textbf{46.1}                          & 75.3 & \textbf{76.4}                  \\
MAGBERT  & 46.9 & 47.1                          & 83.1 & \textbf{83.6}                  \\
MAGXLNET & 46.7 & 47.1                          & 83.9 & 84.2                           \\
MIM      & 53.3 & \textbf{54.1}                          & 79.1 & \textbf{80.1}                  \\
\hline                         
\end{tabular}
\caption{Results on sentiment analysis on CMU-MOSEI. $Acc_{7}$ and $Acc_{2}$ denote accuracy on 7 and 2 classes respectively. Boldface denotes statistical significance. \label{tab:mosei}}
\end{table}



We evaluate our change on the aforementioned five models on four metrics commonly used in the literature \citep{zadeh2017tensor, han2021improving}. The binary and 7-class accuracy results are reported in Tables \ref{tab:mosi} and \ref{tab:mosei}. We present the results of the remaining metrics (MAE and CORR) in Appendix \ref{tab:apx:affect}. We observe consistent improvements in accuracy of non-transformer based models (FLSTM, LFM, MIM) ranging from 0.5\% to 1.5\%, while transformer based models improve by 0.3\%. The comparatively smaller improvement in transformers could be due to the lack of additional information from other modalities when using BERT on text. For example, on CMU-MOSI, simply using BERT embeddings provides an accuracy of 78\% which is higher than most non-BERT fusion models \citep{hazarika2020misa}. Given the degree of  sufficiency in the textual modality, performance is determined by the text network not by the fusion design. 

\subsection{Financial Data}
\textbf{Datasets.}
We evaluate our approach on a multimodal financial time series prediction task \citep{sardelich2018multimodal}. F\&B, HEALTH, and TECH are prices and events related to publically listed companies organized according to the primary business sector. Within each sector, historical prices are used as time series inputs to predict the future price and volatility of a related stock. In this setting the different stocks in the same sector correspond to different modalities. Due to the significantly large number of available modalities, this task presents a different set of challenges \citep{emerson2019trends, sardelich2018multimodal} than other datasets. Moreover, due to the inherently low signal-to-noise ratio in such time series, it presents a greater robustness challenge than other datasets \citep{liang2021multibench}. On the other hand, due to the similar nature of the modalities this task is amenable to early fusion methods.

\textbf{Models.}
We experiment with  Transformers for time series \citep{sardelich2018multimodal} with both early fusion \textbf{EF transf} and late fusion \textbf{LF transf} variants. The other models we test are the multimodal fusion transformer \textbf{MulT}  \citet{tsai2019multimodal}, Gradient Blending \textbf{GB} approach from \citep{wang2020makes}. Finally as LSTMs are strong baselines on this task \citep{eflstm}, we also use Early fusion \textbf{EFLSTM} and Late \textbf{LFLSTM} fusion LSTM models.

\begin{table}[thb]
\centering
\small
\begin{tabular}{|l|l|ll|ll|} 
\hline
\begin{tabular}[c]{@{}l@{}}\\\\\end{tabular} & Metric  & \multicolumn{2}{c|}{MSE \; $\downarrow$} & \multicolumn{2}{c|}{Robustness~ $\uparrow$}  \\ 
\hline
Model                                        & Dataset & Base  & Ours                          & Base & Ours                                   \\ 
\hline
\multirow{3}{*}{EFLSTM}                      & F\&B      & 0.73  & \textbf{0.70}                & 0.87 & \textbf{1.0}                                   \\
                                             & HEALTH  & 0.308 & 0.306                        & 0.54 & \textbf{0.83}                                  \\
                                             & TECH    & 0.742 & \textbf{0.738}               & 0.92 & 0.93                                  \\ 
\hline
\multirow{3}{*}{LFLSTM}                      & F\&B      & 0.77  & \textbf{0.73 }               & 0.74 & \textbf{0.83}                                  \\
                                             & HEALTH  & 0.331 & \textbf{0.315 }              & 0.48 & \textbf{0.78}                                  \\
                                             & TECH    & 0.736 & 0.737                        & 0.96 & 0.96                                  \\ 
\hline
\multirow{3}{*}{GB}                          & F\&B      & 0.690 & 0.688                        & 0.98 & 0.98                                  \\
                                             & HEALTH  & 0.318 & \textbf{0.305 }              & 0.67 & \textbf{1.0}                                   \\
                                             & TECH    & 0.740 & \textbf{0.728 }              & 0.99 & 1.0                                   \\ 
\hline
\multirow{3}{*}{LF Transf}                   & F\&B      & 0.838 & \textbf{0.788}               & 0.24 & \textbf{0.38}                                  \\
                                             & HEALTH  & 0.337 & 0.331                        & 0.34 & \textbf{0.46}                                  \\
                                             & TECH    & 0.757 & 0.751                        & 0.92 & 0.93                                  \\ 
\hline
\multirow{3}{*}{MulT}                        & F\&B      & 0.814 & \textbf{0.765 }              & 0.33 & \textbf{0.48}                                  \\
                                             & HEALTH  & 0.333 & \textbf{0.329 }              & 0.0  & 0.08                                  \\
                                             & TECH    & 0.763 & \textbf{0.757 }              & 0.85 & 0.86                                  \\ 
\hline
\multirow{3}{*}{EF Transf}                   & F\&B      & 0.836 & \textbf{0.827}               & 0.0  & 0.05                                  \\
                                             & HEALTH  & 0.335 & \textbf{0.326 }              & 0.45 & \textbf{0.63}                                  \\
                                             & TECH    & 0.755 & \textbf{0.750 }              & 0.0  & 0.0                                   \\
\hline
\end{tabular}
\caption{Results on stock prediction on the three sectoral datasets. The performance is evaluated on the Mean Squared Error (MSE) metric evaluated on 10 trials. We also compute robustness metrics by testing on data corrupted with various noise levels and present the relative robustness scores. Scores which are outside the 1 standard deviation of the base model are highlighted. \label{tab:finance}}
\end{table}

Because of the similar nature of the modalities one might expect early fusion based models to be effective. This can be seen in our results where early fusion LSTM outperforms late fusion models. However, we note that, by using backward connections, the late fusion models, especially LFLSTM, become competitive with early fusion models. 
The nature of the dataset---low dimension time series with inherent noise---means we can also assess the robustness of the models against modality corruption easily. We add varying levels of noise to the test data and see how the performance of the models change with increasing noise. Following \citep{taori2020measuring} the robustness of the model is assessed by computing the area under the performance vs noise curve. Specifically it is computed by discrete approximation of the following integral:
$$ \tau = \int \text{MSE}(f, \sigma) - \text{MSE}(b, \sigma) d\sigma$$
where $\text{MSE}(.,\sigma)$ is the MSE of the model on test-data with inputs corrupted with noise level $\sigma$. $f$ is the model the evaluated and $b$ is a baseline model. We choose late fusion transformer as our baseline, and scale the numbers to $\left[0,1\right]$.
From the results we can see that \emph{Pro-Fusion provides greater improvements on late fusion compared to early fusion designs}. This suggests that part of the improvement is due to backward connection-based fusion acting as a bridge between early and late fusion designs.


\subsection{Ablation Experiments}

\begin{minipage}[c]{0.45\textwidth}
\centering
\includegraphics[width=0.95\textwidth]{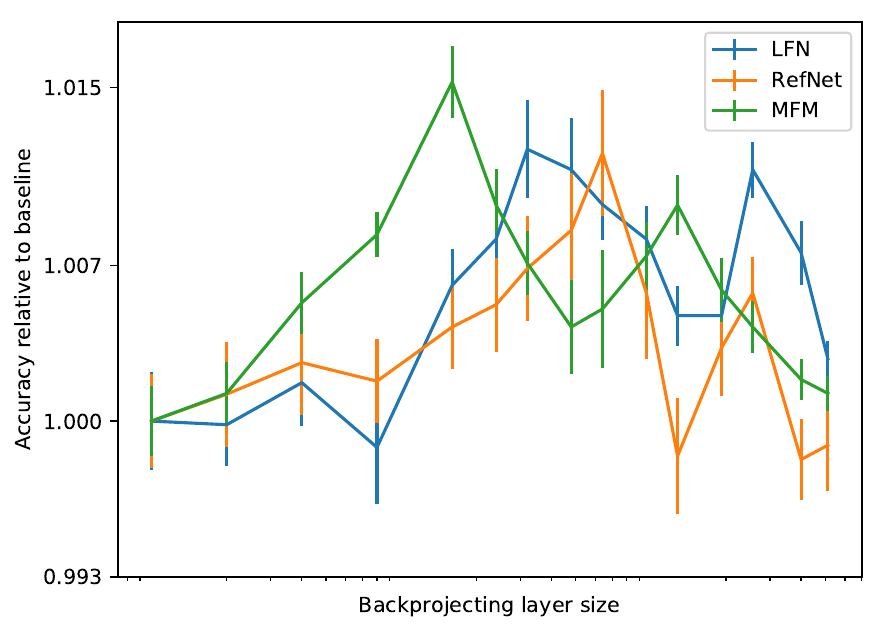}
\captionof{figure}{ Relative accuracy of different models over varying dimensions of backprojecting connections. Each point corresponds to the normalized performance of the corresponding model when the hidden dimension is set to the values of on the x axis.
\label{fig:apx:avmnist_ablation2}}
\end{minipage}
\begin{minipage}[c]{0.45\textwidth}
\centering
\begin{tabular}{|l|l|l|l|}
\hline
 & \multicolumn{3}{c|}{Accuracy $\uparrow$} \\
\hline
Model  &  Base &  Ours & Iterative\\
\hline
\hline
LFN    & 71.1     &   \textbf{71.8}      &  71.5 \\
MFM    & 71.4     &  \textbf{72.2}  &   69.9    \\
GB   & 68.9     &   69.3            &  69.2 \\
Refnet & 70.6    &  \textbf{71.2}    & 70.7      \\
\hline
\end{tabular}
\captionof{table}{Results on digit classification task with AVMNIST for various fusion architectures. The performance metric is Accuracy, and was measured on five trials. \label{tab:apx:avmnist_ablation}}
\end{minipage}

To assess the impact of multimodal backprojecting connections in the Pro-Fusion approach against vanilla iterative models, we conduct experiments on AVMNIST. We change the unimodal feature generators of the baseline models into an iterative model. Effectively, these models are similar to the Pro-Fusion model except that we connect the output features of the unimodal feature generators to their inputs instead of having multimodal connections (See Figure \ref{fig:apx:ablation_unroll} in the Appendix). This allows us to distinguish between the effect of multimodal backprojection from the effect of generic iterative processing. We fixed the number of iterative steps to 2 (same as our Pro-Fusion models) and ran 8 trials for these alternate iterative model. Our results along with baseline and Pro-Fusion models are reported in Table \ref{tab:apx:avmnist_ablation}.

The results indicate that, while iterative models do lead generally to some improvement over the baseline models, Pro-Fusion is still better. Moreover in some cases (such as MFM) iterative models can be worse than the baseline. The key difference between a vanilla iterative model and Pro-Fusion is that Pro-Fusion allows unimodal feature generators access to information from other modalities. As such, unimodal feature generators can now produce  features conditioned on the other modalities. On the other hand, in the alternate approach, the unimodal features are blind to the features from alternate modalities. 

We also run experiments to evaluate the effect of the dimensionality of the backprojecting connections. We adjust the dimensionality of the backprojecting connection $W$, up to 512 and evaluate multiple models on  AVMNIST. One might expect that backprojections with very low dimensions will be similar to baseline models with no backward connection. On the other hand, with a high dimensionality in the backward connection, one runs into the same problem as early fusion of high parametric complexity. This expectation matches the empirical results, as can be seen in Figure \ref{fig:apx:avmnist_ablation2}, where we plot the accuracy (and standard error) of multiple models with varying backprojection sizes. Notice that, for comparability across models, we have normalized all curves with their respective baseline results.

\section{Conclusion}
Our paper presents a model-agnostic approach to incorporate benefits of early fusion into late fusion networks via backward connections. We argued for some sufficient conditions when our backward connection based design to be more effective than usual fusion designs, supported by an artificial data experiment. With real data experiments, we make a case for using multimodal backward connections and show that Pro-fusion can improve even SOTA models.

\textbf{Limitations}: Our experiments do not give a clear answer to the question of how our approach interacts with other model agnostic methods. As Pro-Fusion instantiates a version of message passing on the graph in Figure \ref{fig:M1}, if a task has a similar dependency structure, then our method can be expected to yield good improvements \footnote{for details refers to Appendix \ref{apx:ablation2}}. That said, no model design is suitable for all problems, and our results might not generalize to other datasets.

\bibliography{mybib}

\clearpage
\onecolumn
\setcounter{thm}{0}

\appendix

\section{Overview of Fusion Techniques and Related Works}
\label{apx:related}
Multimodal fusion has been a heavily researched area for decades \citep{Osadciw2009,varshney1997distributed}. Such models have been used for tasks ranging from video classification  \citep{yang2016multilayer}, action recognition \citep{cabrera2019gestures}, and speech enhancement  \citep{hou2017audio} to brain studies \cite{sui2012review}, ecological applications \citep{taheri2018animal}, and monitoring systems \citep{varshney1997distributed,li2018driver}. However most models have primarily focused either on architectural changes or designing new fusion layers \citep{yan2021deep}.

\paragraph{Early Fusion} Figure \ref{fig:base_fusion:early} illustrates a general early fusion scheme. Early fusion, sometimes also called feature fusion in the early literature \citep{ayache6,chair1986optimal}, creates a  multimodal representation by combining unimodal information before they are processed. One can broadly interpret an Early Fusion scheme as one that integrates unimodal features before `learning high level concepts'. Early fusion models have the ability to model highly complex dependencies between different modalities, however they generally face problems when dealing with heterogenous sources such as text and images.

\paragraph{Late Fusion} In contrast to early fusion, late fusion designs learn `high level semantic concepts' directly from unimodal features. Late fusion allows an easy way to aggregate information from diverse modalities and can easily incorporate pre-trained models (e.g. \citet{rahman2020integrating}). As such, late fusion has been the more commonly utilized framework for multimodal learning \citep{ramachandram2017deep,simonyan2015two}. However, late fusion models, illustrated in Figure \ref{fig:base_fusion:late}, run the risk of missing cross-modal interactions in the mixed feature space.

\paragraph{Architecture Changes}
Due to the wide variety of applications and tasks which require multimodal fusion, over the years a plethora of different architectures have been used. Some of the recent works include that of \citet{vielzeuf2018centralnet}, \citet{sankaran2021multimodal}, \citet{perez2019mfas}, \citet{hazarika2020misa}, and \citet{khan2012color}. \citet{vielzeuf2018centralnet} proposed a multimodal fusion design called CentralNet that is based on aggregative multi-task learning. \citet{sankaran2021multimodal} bring together ideas from CentralNet and Cycle-GAN \citep{zhu2017unpaired} and design a Refiner Fusion Network (Refnet). The Refnet design uses a de-fusion model trained via cyclic losses to align both unimodal and multimodal representations in a common latent space. \citet{perez2019mfas} used neural architecture search to find a good architecture for convolutional networks. The discovered architecture is a multistep fusion model that fuses information from different individual unimodal layers multiple times. \citet{hsu2018disentangling} and \citet{khattar2019mvae} used ideas from unsupervised learning to use multimodal autoencoders to learn better representations. \citet{tsai2019multimodallearning} improved upon the factor model based approach of \citet{hsu2018disentangling} by incorporating prior matching and discriminative losses. \citet{nagrani2021attention}  modify the multimodal transformer \citet{tsai2019multimodal} to incorporate bottlenecks.

Our proposed method, though technically an architecture change, is a single change that \emph{treats the existing model as given}. It is closer in spirit to a black-box change, compared to the aforementioned methods. Hence it is \emph{complementary} to this line of work. We experiment with many of the aforementioned models to show how our proposal consistently improves performance.

\paragraph{Fusion Techniques}
Other than basic fusion layers such as pooling and concatenation, other common layers use include aggregation \citep{khan2012color}, tensor factorisation \citep{liu2018efficient}, attention modules \citep{zadeh2018memory, tsai2019multimodal}, channel-swaps \cite{wang2020cen} and non-local gating \citep{hu2019squeezeandexcitation,wang2018non,liu2019contextualized}. \citet{rahman2020integrating} used pre-trained transformer \citep{siriwardhana2020jointly} along with \citet{wang2019words} modulation gate to achieve state of the art results on the multimodal sentiment benchmarks MOSI \cite{wollmer2013youtube} and MOSEI \cite{zadeh2018multimodal}. LFN \citep{zadeh2017tensor} combined information via pooling projections of high dimensional tensor representation of multimodal features. 
These works propose specific fusion techniques, they design specific forms of the $F$ function (see Figure \ref{fig:base_fusion}). Our proposed technique is \emph{agnostic to the choice of the fusion function $F$} and thus is \emph{orthogonal} to these ideas. 

\paragraph{Model Agnostic Methods}
A number of alignment and information based losses have also been explored to improve fusion by inducing semantic relationships across the different unimodal representations \citep{abavisani2019improving, bramon2011multimodal, Liu_2021_ICCV, han2021improving} \nocite{shankar2022multimodal}. These are purely train-time objectives and can be generally applied to most multimodal fusion models.
Recently \citet{wang2020makes} proposed a new approach that can be applied to any multimodal architecture. Their approach called Gradient Blending (GB) tackles the problem of joint learning when different unimodal networks have varying capacity, by learning individual modality weights factors based on the model performances. Our proposal instead of adding losses or adding reweighing factors instead adds backprojective connections. So, these model-agnostic proposals are in general complementary to our approach, and can be \emph{combined with it to achieve further improvements}. 


\paragraph{Memory based Fusion}
Existing works using multiple fused representation computed over time, have been used for sentiment analysis \citep{gammulle2017two,zadeh2018memory,zadeh2018multi}. The purpose of memory in those methods is retaining history for easier learning of interactions across time steps in a sequential input. On the other hand, in our proposal, context vector serves the purpose of making late-fusion features accessible to the unimodal network processing models/early stage features. Our proposal is entirely independent of any temporal axis/sequential nature in the input.  Secondly, these methods capture historical relationships only in unimodal data, and memory is used directly over the concatenated unimodal features. On the other hand our approach provides multimodal information to unimodal feature generators.

\paragraph{Deep Unfolding} Iterative neural networks \citep{chang2000structural,hershey2014deep} have been successful for a variety of problems in computer vision such as inverse problems \citep{adler2017solving,chun2020momentum}, super resolution \citep{neshatpour2019icnn} and other tasks \citep{chang2000structural,balatsoukas2019deep}. Deep unfolding methods are specific recurrent models which at each iteration, pass the results of inference from previous iterations onward. Most such methods have been used for image super resolution \citep{zhang2020deep,ning2020accurate} and wireless communication systems \citep{balatsoukas2019deep}
While Pro-Fusion shares an unrolling design with these works, it differs from these methods in the following ways:

\begin{itemize}
\item Unrolling is used primarily as a way to train the backward connections, and are not the fundamental aspect of Pro-Fusion. ProFusion was specifically designed to solve a problem in multimodal data fusion: the “fuse it or lose it” situation; by adding cross-modal backward connections. In principle other methods such as equilibrium propagation \citep{ernoult2020equilibrium}, balance-tuning \citep{zhang2018brain} or other methods can also be used to train a self-iterative loop of Pro-Fusion models.

\item Deep unfolding was not designed for multimodal data, nor adapted to it, to the best of our knowledge. Deep unfolding/iterative models that do not cycle cross-modal information still suffer from the “fuse it or lose it” problem. On the other hand, with ProFusion, unimodal representations adapt to multimodal features, as the model can, in future iterations, extract complementary information based on previous cross-modal features.

\item Deep unfolding methods unroll a classical iterative algorithm such as “gradient descent” or “orthogonal message passing” and introduce trainable parameters for the update step. On the other hand, we consider a given multimodal fusion model as one step of the iteration.

\item In our approach, parameters are shared between steps, which is not common in deep unfolding literature, where every iteration typically introduces a new set of parameters.
\end{itemize}

\section{Further Experiments}


\subsection{Synthetic Experiments}
\label{apx:synth}
We further explore the setting implied by the generative model described in Section \ref{sec:motiv}. For this we generate data as from a generative model matching the dependencies in Figure \ref{fig:M1}. We set the function $h$ to be leaky-Relu and $g$ to be the sine function. We choose $f$ such that $f \circ h$ is linear. Note that the specific choice of $h,g$ makes the function $h+g$ non-invertible. $Z$ was sampled from a uniform distribution on [-2.5,2.5] and all linear transform matrix were also sampled from the standard normal distribution. The specific generative equations are presented in the equations below.

\begin{minipage}{0.5\textwidth}
\includegraphics[width=0.7\linewidth]{newfigs/Recurrent_Fusion_graphical_model.pptx.jpg}
\captionof{figure}{
\label{fig:synth_cont}}
\end{minipage}
\begin{minipage}{0.5\textwidth}
\begin{align}\label{eA1}
& Z \sim \mathcal{U}[-2.5,2.5]^d\\
    &W_1 \sim \mathcal{N}(0,I), W_2 \sim \mathcal{N}(0,I), W_y \sim \mathcal{N}(0,I) \\
    & X_1 = \text{lRelu}(W_1Z) - 2*|\eta|\sin(W_2Z) + \epsilon_1 \\
    & X_2 = \sin(W_2Z) + \sigma_2 \epsilon_2 \\
    & Y = W_yZ + \epsilon_y\\
\end{align}\null
\par\xdef\tpd{\the\prevdepth}
\end{minipage}

\begin{minipage}{0.5\textwidth}
\captionsetup[figure]{width=.8\textwidth}
\includegraphics[width=\linewidth,keepaspectratio=true]{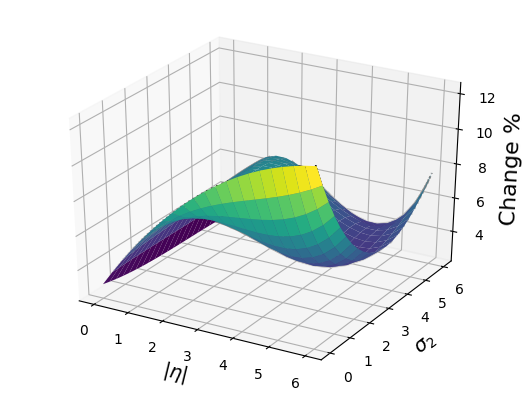}
\captionof{figure}{Percentage Improvement over varying levels of modality dependence. Note that this is percent improvement in MSE so higher is better
\label{fig:synth_cont_seed}}
\end{minipage}
\begin{minipage}{0.5\textwidth}
\includegraphics[width=0.9\linewidth,keepaspectratio=true]{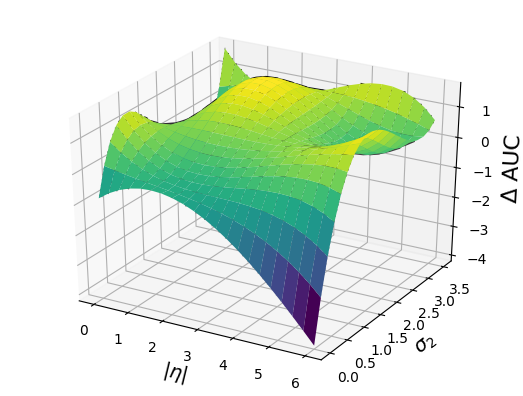}
\captionof{figure}{AUC change for pro-fusion model over normal fusion over varying levels of modality dependence. Since the metric is error, lower AUC is better
\label{fig:synth_rob}}
\end{minipage}

All noise terms $\epsilon$ are sampled from a standard normal distribution. We train a 3 layer MLP to solve this regression task. Fusion was done via plain concatenation of the second layer. For the pro-fusion model, the same fusion vector was linearly transformed and fed along with the input. We varied the strength $\eta$ of corruption in $X_1$ and of $\sigma_2$ the noise in $X_2$, and ran for each such value 30 trials. In close to \textit{90}\% of all trials we found the pro-fusion model to perform better with an average improvement of \textit{8}\%. We plot the contours for multiple different runs in Figures \ref{fig:synth_cont_seed}.

We also do robustness evaluation of this model over different values of noise parameters. We use the AUC (area under ROC curve) metric for this purpose. In Figure \ref{fig:synth_rob} we plot the improvement in AUC of the pro-fusion model over direct fusion against the different values of $\eta/\sigma_2$

\subsection{Exploratory Experiments}

In this section we explore various aspects of the backprojecting connections, such as the layer at which backprojecting connection joins, the number of unrolling steps and the inference complexity.  

\begin{figure}[htb]
    \centering
    \begin{subfigure}[b]{0.45\textwidth}
    \includegraphics[width=0.95\textwidth]{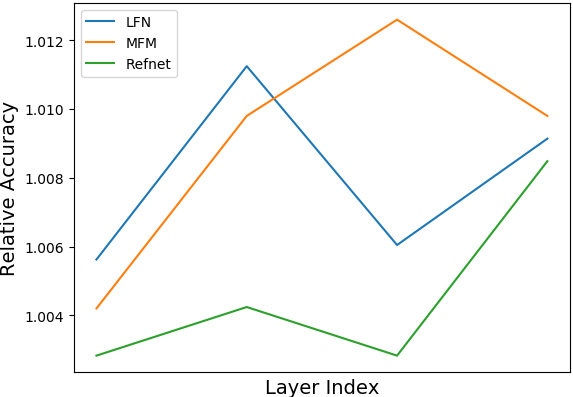}
    \caption{Performance against layer index of backprojecting connection}
    \label{fig:accuracy_across_layers}
    \end{subfigure}
    \begin{subfigure}[b]{0.45\textwidth}
     \centering
     \includegraphics[width=0.92\textwidth]{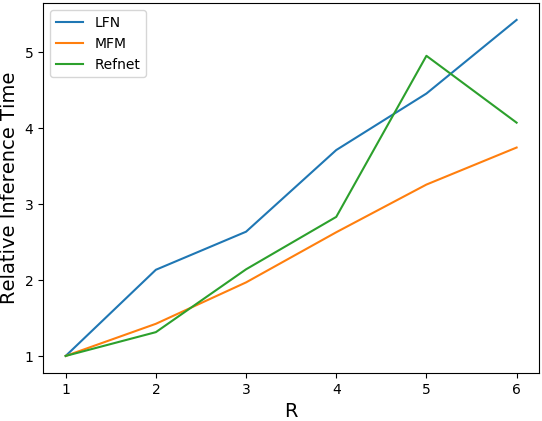}
     \caption{Inference time vs R    \label{fig:my_label1}}
     \end{subfigure}
\end{figure}

We measure the training time and inference time for pro-fusion with different architectures relative to that for the base model for different number of unrolling steps $R$. The straightforward way in which the current pro-fusion design uses the base model, suggests that both training and inference time should vary proportionately to the number of steps. This expectation is brought out in our experiments and can be seen in Figure \ref{fig:my_label1}.

We also conduct experiments with determining at which upstream layer should the fused out should be connected back to. For this once again we run trials on the AVMNIST data. Our results are depicted in Figure \ref{fig:accuracy_across_layers}.

\section{Financial Data Details}

The data itself is not licensed but following \citet{liang2021multibench,sardelich2018multimodal} can be gathered from online records of historical stock prices and events.

\begin{itemize}
    \item \textbf{FB} is composed of S\&P 500 stocks which are part of food, meats and restaurant chains and includes the tickers CAG, CPB, DRI, GIS, HRL, HSY, K, MCD, MKC, SBUX, SJM, TSN, YUM.
    \item \textbf{HEALTH} is composed of following health-care and pharmaceutical sector tickers MRK, WST, CVS, MCK, ABT, UNH, TFX, PFE, GSK, NVS, WBA.
    \item \textbf{TECH} is composed of technology and information service stocks from NASDAQ. We include tickers AAPL, ADBE, AMD, AMZN, GOOG, HPQ, IBM, INTC, MSFT, MSI, NVDA, ORCL, QCOM, ZBRA.
\end{itemize}

\label{apx:fin_robust}

\subsection{Robustness Computation}
We use the area under the performance-noise curve as the measure of robustness. This measure of robustness is the same as the one used by \citet{liang2021multibench, taori2020measuring}, and can be computed via the following integral:
$$ \text{ROBUSTAUC} = \tau = \int Perf(f, \sigma) - Perf(b, \sigma) d\sigma$$
where $Perf(.,\sigma)$ is the performance metric evaluated on a dataset corrupted with noise level $\sigma$, $f$ is the model to be evaluated and $b$ is a baseline model. Basically, the model is evaluated on the same dataset corrupted on an equally spaced grid of noise levels and the performance is averaged over all the noise configurations is used. Note that the $Perf$ as used by \citet{shankar2017no, taori2020measuring} is a positive metric like accuracy. For inverse metrics like MSE one has to use the negative of the above integral. For computing robustness we our experiments we use a transformer model as the baseline.

\section{Experimental Details}
\label{apx:comp_res}
In this section we present the total results of all the experiments. We include metrics such as CORR (pearson correlation) which were not reported in the main body. We also report the average deviation of the scores in these tables.

\subsection{AUC Measures and Average deviation on Financial Data}
\begin{table*}[!htb]
\centering
\small
\makebox[\textwidth][c]{
\begin{tabular}{|l|l|ll|ll|ll|}
\hline
    &
                & \multicolumn{2}{c|}{F\&B} & \multicolumn{2}{c|}{HEALTH} & \multicolumn{2}{c|}{TECH} \\
                \hline
Model &      & MSE  $\downarrow$        & ROBUSTAUC $\uparrow$         & MSE $\downarrow$        &  ROBUSTAUC $\uparrow$         & MSE $\downarrow$  &  ROBUSTAUC $\uparrow$        \\
\hline 
\multirow{2}{*}{EFLSTM} & Base       & 0.73 (0.03)          &  0.35        & 0.308 (0.005)           &  0.018         & 0.742 (0.006)          &   0.027       \\
 &Our        & 0.70 (0.04)          &   0.40      & 0.306 (0.003)           &    0.029       & 0.738 (0.004)          &    0.028      \\

\multirow{2}{*}{LFLSTM} & Base       & 0.77 (0.05)          &    0.29      & 0.331 (0.009)           &    0.016       & 0.736 (0.006)          &   0.028       \\
 &Our        & 0.73 (0.05)          &   0.33       & 0.315 (0.007)           &        0.026   & 0.737 (0.005)          &    0.028      \\
\multirow{2}{*}{GB}  & Base           & 0.690 (0.04)          &       0.39   & 0.318 (0.03)           &    0.022      & 0.740 (0.006)         &   0.029       \\
 & Our        & 0.688 (0.02)          &    0.39      & 0.305 (0.003)             &     0.035      & 0.738 (0.005)         &  0.029    \\
 \multirow{2}{*}{LF Transformer} & Base &     0.838 (0.004)          &       0.09   &    0.337 (0.004)            &   0.012      &       0.757 (0.005)         &   0.027       \\
 & Our &     0.788  (0.004)  &    0.15  &  0.331 (0.004)        &          0.015    &        0.755 (0.005)                  &      0.028    \\
\multirow{2}{*}{MulT}  & Base          & 0.814 (0.005)          &       0.13   & 0.333 (0.004)           &   0.001        & 0.763 (0.005)         &     0.025     \\
 &Our        & 0.765 (0.006)          &    0.19      & 0.329 (0.005)      &   0.002    &     0.757 (0.004)       &    0.026       \\
\multirow{2}{*}{EF transformer} &Base            & 0.836 (0.009)          &    0.      &  0.335 (0.001)           &   0.       & 0.755 (0.004)         &   0.        \\
 & Our        & 0.827 (0.009)          &   0.01       &  0.326 (0.004)             &   0.02        & 0.750 (0.004)         & 0.00  \\
\hline
\end{tabular}
}
\caption{Results on stock prediction on the three sectoral datasets. The performance is evaluated on the Mean Sqaured Error (MSE) and ROBUSTAUC metric. Note we have already flipped the AUC sign for the inverse metric. \label{tab:finance1}}
\end{table*}

Our results on financial time-series prediction, while qualitatively similar to results of \citep{liang2021multibench}, is different because of using more number of target stocks and different time period. For completeness we also report the error on the dataset provided by them in Table \ref{results:finance_supp}.

\begin{table*}[]
\fontsize{9}{11}\selectfont
\setlength\tabcolsep{6.0pt}
\caption{Results on multimodal dataset of \citet{liang2021multibench} in the finance domain}
\centering
\footnotesize
\begin{tabular}{|l|cc|cc|cc|}
\hline
Dataset & \multicolumn{2}{c|}{\textsc{F\&B}} & \multicolumn{2}{c|}{\textsc{Health}} & \multicolumn{2}{c|}{\textsc{Tech}} \\
Metric & \multicolumn{2}{c|}{MSE $\downarrow$} & \multicolumn{2}{c|}{MSE $\downarrow$} & \multicolumn{2}{c|}{MSE $\downarrow$} \\
\hline
 & Base & Our & Base & Our & Base & Our \\
\hline

EF-LSTM & 1.836 & 1.753 & 0.521 & 0.511 & 0.119 & 0.124 \\
LF-LSTM & 1.891 & 1.786 & 0.545 & 0.522 & 0.120 & 0.121 \\
LF-Transformer & 2.157 & 2.112 & 0.572 & 0.566 & 0.143 & 0.144 \\
MulT & 2.056 & 2.032 & 0.554 & 0.553 & 0.135 & 0.132 \\
\hline

\end{tabular}

\label{results:finance_supp}
\end{table*}


\begin{table*}[ht]
\centering
\begin{subtable}{0.5\linewidth}
\centering
\begin{tabular}{|l|llll|}
\hline
     &  $Acc_{7} \uparrow$   & $Acc_{2} \uparrow$   & MAE $\downarrow$  & CORR $\uparrow$\\
\hline
\hline
   & & \multicolumn{2}{c}{FLSTM} &  \\
\hline
Base & 31.2 (0.5) & 75.9 (0.5) & 1.01 & 0.64 \\
Our & \textbf{31.8} (0.4) & \textbf{76.8} (0.3) & 1.0 & 0.66 \\

\hline
   & & \multicolumn{2}{c}{LFN} &  \\
\hline
Base & 31.2 (0.4) & 76.6 (0.4) & 1.01 & 0.62 \\
Our & \textbf{32.1} (0.6) & \textbf{77.2} (0.2) & 1.01 & 0.62 \\

\hline
   & &\multicolumn{2}{c}{MAFBERT} & \\
\hline
Base & 40.2 (0.4) & 83.7 (0.3) & 0.79 & 0.80 \\
Our & \textbf{40.8} (0.4) & \textbf{84.1} (0.3) & 0.79 & 0.80 \\

\hline
   & &\multicolumn{2}{c}{MAGXLNET} & \\
\hline
Base & 43.1 (0.2) & 85.2 (0.4) & 0.76 & 0.82 \\
Our & \textbf{43.5} (0.3) & 85.5 (0.2) & 0.76 & 0.83 \\

\hline
   & &\multicolumn{2}{c}{MIM} & \\
\hline
Base & 45.5 (0.1) & 81.7 (0.2) & 0.72 & 0.75 \\
Our & \textbf{46.3} (0.2) & \textbf{83.4} (0.5) & 0.71 & 0.77 \\

\hline
\end{tabular}
\end{subtable}
\begin{subtable}{0.5\linewidth}
\centering
\begin{tabular}{|l|llll|}

\hline
   &  $Acc_{7} \uparrow$   & $Acc_{2} \uparrow$   & MAE $\downarrow$  & CORR $\uparrow$\\
\hline
\hline
   & & \multicolumn{2}{c}{FLSTM} &  \\
\hline
Base & 44.1 (0.2) & 75.1 (0.3) & 0.72 & 0.51\\
Our &  \textbf{44.8} (0.5) & \textbf{75.8} (0.3) & 0.72 & 0.52\\
\hline
   & & \multicolumn{2}{c}{LFN} & \\
\hline
Base &  44.9 (0.3) & 75.3 (0.4) & 0.72 & 0.52 \\
Our &  \textbf{46.1} (0.3) & \textbf{76.4} (0.3) & 0.71 & 0.52 \\

\hline
   & & \multicolumn{2}{c}{MAFBERT} & \\
\hline
Base &  46.9 (0.7) & 83.1 (0.4) & 0.59 & 0.76 \\
Our &  47.1  (0.7) & \textbf{83.6} (0.2) & 0.58 & 0.77 \\

\hline
   & & \multicolumn{2}{c}{MAGXLNET} & \\
\hline
Base &  46.7 (0.4) & 83.9 (0.3) & 0.59 & 0.77 \\
Our &  47.1 (0.3) & 84.2 (0.3) & 0.57 & 0.77 \\

\hline
   & & \multicolumn{2}{c}{MIM} & \\
\hline
Base &  53.3 (0.5) & 79.1 (0.3) & 0.59 & 0.71 \\
Our &  \textbf{54.1} (0.8)  & \textbf{80.1} (0.2) & 0.57 & 0.73 \\

\hline
\end{tabular}
\end{subtable}
\caption{Results on sentiment analysis on a) CMU-MOSI and b) CMU-MOSEI. $Acc_{7}, Acc_{2}$ denote accuracy on 7, 2 classes respectively. $MAE$ is Mean Absolute Error and Corr is the Pearson correlation. \label{tab:apx:affect}}
\end{table*}

\subsection{Multimedia}
Complete results on AVMNIST along with the standard deviations of the performance are reported in Table \ref{tab:apx:avmnist}
\begin{table}[]
\centering
\begin{tabular}{|l|l|l|}
\hline
 & \multicolumn{2}{c|}{Accuracy $\uparrow$} \\
\hline
Model  &  Base &  Ours \\
\hline
\hline
LF     & 71.4 (0.4)    &   71.6 (0.4)           \\
LFN    & 71.1 (0.3)    &   \textbf{71.8} (0.3)        \\
MFM    & 71.4 (0.4)    &  \textbf{72.2} (0.6)        \\
GB   & 68.9 (0.6)    &   69.3 (0.5)                 \\
Refnet & 70.6 (0.7)    &  \textbf{71.2} (0.5)          \\
MFAS & 72.1 (0.5)    &  \textbf{72.5} (0.3)          \\
\hline
\end{tabular}
\caption{Results on digit classification task with AVMNIST for various fusion architectures. The performance metric is Accuracy. Scores outside the average range of baseline models have been highlighted. \label{tab:apx:avmnist}}
\end{table}

\subsection{Hyperparameter Details}
For the AVMNIST dataset, we used LeNet style unimodal feature generators. For the image encoder we used a 4 layer network with filter sizes [5,3,3,3] and max-pooling with width of 2. For the audio encoder the networks was a
6 layer networks with filter sizes [5,3,3,3,3,3] and max-pooling of width 2. The channel width was doubled after each layer. For GB models, the validation size was 0.8 and the model is fine-tuned for gradient blending for 30 epochs. For the optimization process we tried random search on a logarithmic scale on the interval [1e-5, 5e-2]. We experimented with Adam, Adagrad, RMSProp, SGD optimizer with default configurations.  

\begin{figure}[htb]
    \centering
    \hspace{1cm}
    \includegraphics[width=0.55\textwidth]{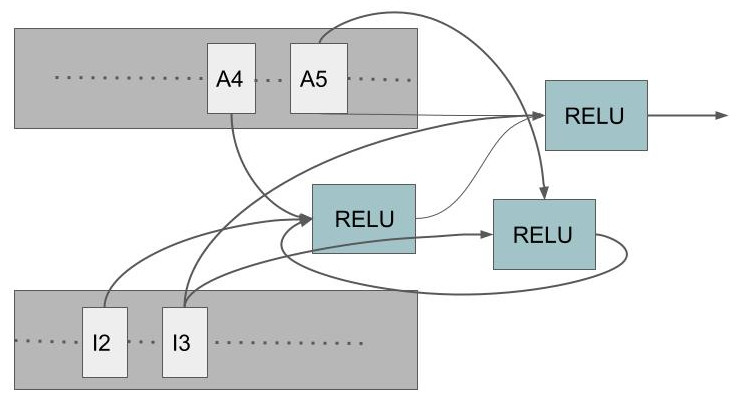}
    \caption{MFAS \citep{perez2019mfas} based Multimodal Fusion Architecture for AVMNIST. Every arrow into the activation corresponds to a linear layer. The A4 and A5 represent the fourth and fifth layer of the audio encoder. Similarly I2 and I3 represent the second and third layer of the image encoder.}
    \label{fig:mfas_mnist_design}
\end{figure}

For the MFAS model, we did not do architecture search but instead used the final model presented by \citet{perez2019mfas}. That model is shows in Figure \ref{fig:mfas_mnist_design}. While we have tried to stay close to the method described in \citet{perez2019mfas, liang2021multibench} for creation of this dataset, our version of AVMNIST is potentially different from the earlier reported results as no standard dataset is available. For financial time series prediction, we used 128 dimensional RNNs. For transformers we used a 3 layer network with 3 attention heads. The sequence length used for BPTT in all cases was 750. The optimization process was chosen in a similar way as mentioned previously.

\label{apx:mfas_avmnist}

\paragraph{Parameter Sizes}
\textbf{} \\

\begin{table}[]
\centering
\begin{tabular}{|l||cc|}
\hline
Model  & Base  & Ours  \\
\hline
LFN    & 1626k & 1655k \\
MFM    & 1142k & 1163k \\
RefNet & 582k  & 607k  \\
GB     & 659k  & 671k \\
\hline
\end{tabular}
\caption{Parameter size comparison for models tested on AVMNIST}
\end{table}

Models on MOSI/MOSEI, use and fine-tune BERT (or another similar large language model). The total number of trainable parameters for these models is >20M, and so the additional parameters introduced by ProFusion are especially small (relatively). In the  We will add a table in the appendix for the number of parameters for all models in the final version.

\section{Technical Analysis}
\label{apx:theory}

Given a base model $\mathcal{F}$ with input $x = (x_i,x_2,..x_k)$, we want to create an augmented model $\hat{\mathcal{F}} : \mathcal{X} \times \mathbb{R}^{d} \rightarrow \mathcal{Y}$ with additional input $c \in \mathbb{R}^{d}$ such that $ c = 0 \implies \hat{\mathcal{F}}(x,c) = \mathcal{F}(x)$. Recall that the function $\mathcal{F}$ mentioned in Section \ref{sec:prelim} is given by $
\mathcal{F}(x) = P(F(G_1(x_1),G_2(x_2),..G_K(x_K)))
$. 

We create the desired network $\hat{\mathcal{F}}$ by providing $c$ to the unimodal feature generators $G_j$. We use the output of the fusion layer $F$ and project it back into the network as $c_t$ via the matrix/function $W_i$. Specifically we choose a modified generator $\hat{G}_i : \mathcal{X}_i \times \mathbb{R}^{d_i}$ to be given by $\hat{G}_i(x_i,c) = G_j([x_i, W_i(c)])$, where $W_i$ represents a matrix/network. 
This creates a recurrence relation which we unroll for $R$ steps. The final vector $c_R$ after $R$ steps serves as the output of fusion which is then provided to the predictor model $P$.

Mathematically, we can write the overall operation as :
\begin{align}
\label{eqn:apx_mod}
    & \hat{G}_i(x_i,c_{t-1}) = G_j(x_i + W_i(c_{t-1})) \\
    &c_t =  E(F(\hat{G}_1(x_1,c_{t-1}),..,\hat{G}_K(x_K, c_{t-1}))) \\
    &\hat{Y} = P(c_R) \\
\end{align}
with the initial value $c_0 = \vec{0}$.
We would like to draw the readers attention to the lack of \red{$t$} subscript on the inputs $x$ in Equation \ref{eqn:apx_mod} above. This is because the iterations on $t$ are on a dimension unrelated to any sequentiality in the input. Instead the model iteratively modifies its late-fusion features output, and makes it available to the unimodal network processing models/early stage features via the vector $c_t$. This is  different from models like \citep{zadeh2018memory} which use memory to store fusion vectors across different time-steps in the input, i.e. their networks processes $x_t$ to produce fusion output.


%

\subsection{Linear Model}

Consider a simple linear model with multiple outputs and two input modalities. The two modalities are each $\mathbb{R}^D$ and the number of outputs is $K$. We consider the late and early fusion model of Figure \ref{fig:apx:mult_nl_theory} with only 1 layer between the input and fusion and from fusion to output. We take the fusion layer $F$ to be a concatenating operation. While the figure includes non-linearity, we will in this discussion look at the linear case.

\textbf{Analysis} The early fusion model works by concatenating the two inputs together, a linear transform to $\mathbb{R}^{2d}$ followed by another transform to $\mathbb{R}^{K}$. On the other hand the late fusion model corresponds to two modality wise transform to $\mathbb{R}^{d}$ which are then concatenated and transformed to $\mathbb{R}^{K}$. Note that in either case the second transformation is a $2d \times K$ matrix. 

$$
 \begin{bmatrix}
F_{11} & F_{12} \\
F_{21} & F_{22} 
\end{bmatrix}
 \begin{bmatrix}
W_{11} & W_{12} \\
W_{21} & W_{22} 
\end{bmatrix}
\begin{bmatrix}
X_{1}\\
X_{2}
\end{bmatrix}
=
\begin{bmatrix}
F_{11}W_{11} + F_{12}W_{21} & F_{11}W_{12} + F_{12}W_{22} \\
F_{21}W_{11} + F_{21}W_{22} & F_{21}W_{12} + F_{22}W_{22}
\end{bmatrix}
\begin{bmatrix}
X_{1}\\
X_{2}
\end{bmatrix}
$$
Late fusion on the other hand is expressible in the same formula by zeroing the off diagonal terms of $W$
$$
\begin{bmatrix}
F_{11} & F_{12} \\
F_{21} & F_{22} 
\end{bmatrix}
 \begin{bmatrix}
W_{11} & 0\\
0 & W_{22} 
\end{bmatrix}
\begin{bmatrix}
X_{1}\\
X_{2}
\end{bmatrix}
= 
\begin{bmatrix}
F_{11}W_{11}  &  F_{12}W_{22} \\
F_{21}W_{11} &   F_{22}W_{22}
\end{bmatrix}
\begin{bmatrix}
X_{1}\\
X_{2}
\end{bmatrix}
$$
First note that the rank of the effective matrix in early fusion is necessarily higher than that in late fusion. However even in cases when the rank of the effective matrix remains the same the matrix in late fusion is more constrained. To see this, consider a simple case where all $F_{ij}$ are diagonal matrices; in which case the ratio between the rows in the top-left quadrant (i.e. $F_{11}W_{11}$) is the same as the one in the bottom-left quadrant (i.e. $F_{21}W_{11}$) (and the same is true for the other quadrants as well). Similar constraints hold more generally. For example take the case of $d=1, D=2, K=2$. The net transformation in this case is a $4$x$2$ matrix of the form:

$$
\begin{bmatrix}
f_{11} w_{11} & f_{11} w_{12} & f_{21}w_{21} & f_{21} w_{22} \\
f_{12} w_{11} & f_{12} w_{12} & f_{22}w_{21} & f_{22} w_{22} \\
\end{bmatrix}
$$
One can see that the ratio of the elements of the first column and second column is the same. The same holds for third and fourth columns. Evidently not all rank two 4x2 matrices are of this form. As such there are functions in the early fusion variant which cannot be expressed in the late fusion design.

Next we compare this to the pro-fusion design presented in this work. We use a variant where the concatenated late vector is projected back as an input to the input encoders \ref{fig:base_fusion:rec}, and the entire network is unrolled once.

The composite action of the backward connection plus unrolling is still linear and is given by the following composition
$$
\begin{bmatrix}
F_{11} & F_{12} \\
F_{21} & F_{22} 
\end{bmatrix}
 \begin{bmatrix}
W_{11} + G_{11} W_{11} & G_{12} W_{22}\\
G_{21} W_{11} & W_{22}  + G_{22} W_{22}
\end{bmatrix}
\begin{bmatrix}
X_{1}\\
X_{2}
\end{bmatrix}
$$

The presence of off-diagonal entries which similar to early fusion breaks the structure imposed by late fusion in the earlier case. However this model is not as expressive as early fusion as there can be some dependencies between the matrix entries.

\subsection{Multiplicative Nonlinearity}

\begin{figure*}
    \centering
    \includegraphics[width=0.8\textwidth]{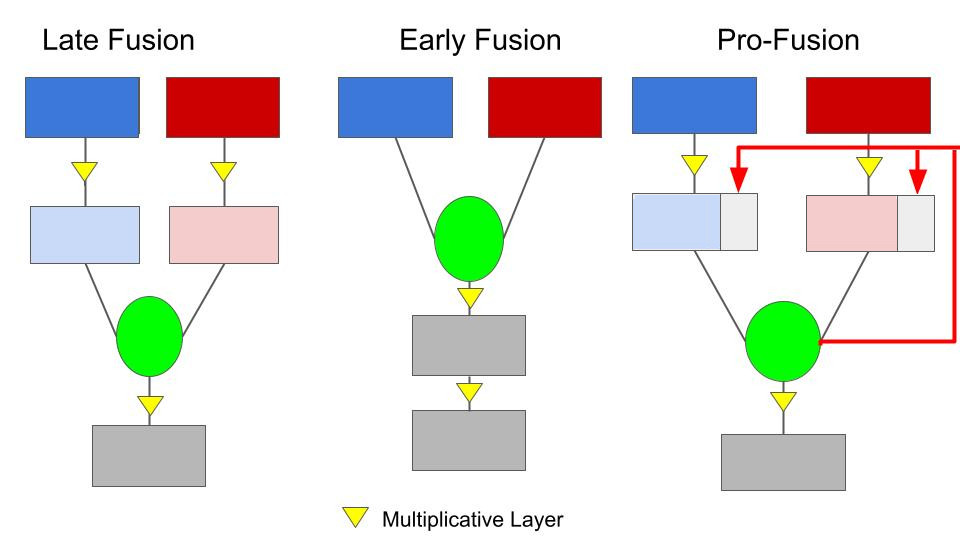}
    \caption{Representative Multimodal Fusion Architectures of Late fusion , Early fusion and Pro-fusion. The round green layer is the fusion layer (assumed to be concatenation). We also depict on the figure the location of multiplicative non-linearity with triangle,  and highlight in red the back-projections of the pro-fusion design}
    \label{fig:apx:mult_nl_theory}
\end{figure*}

While in the linear case, the extra freedom allowed by progressive fusion need not be very useful, the presence of multimodal interaction/ off-diagonal terms can have larger effects when dealing with non-linearity.

Consider a multiplicative non-linear layer $ H : \mathbb{R}^D \rightarrow  \mathbb{R}^d$ with the ability to provide any $d$ features obtained via pairwise multiplication of input features. For example for a vector input $[x_1, x_2, x_3, x_4]$ and output dimension $d=3$, we can get any 3 of the 10 pairwise outputs i.e $[x_1^2, x_2^2, x_3^2, x_4^2, x_1x_2, x_1x_3, x_1x_4, x_2x_3, x_2x_4, x_3x_4 ]$. Using multiplicative non-linearity is useful for analysis as using the distributive property one can directly include behaviour of linear transformations.
In Figure \ref{fig:apx:mult_nl_theory}, we depict three simple non-linear fusion models with such non-linearity added in the layers. We denote the input features as $X^1, X^2$ for the two modalities, and their individual components are denoted by $X^1_1,X^1_2, X^2_1$ etc. We would also refer by $w^{p,q}$ the weight matrix applied on modality $p$ in the layer $q$ of the network.

\paragraph{Analysis}
After the first layer in the late-fusion design, the unimodal features are $\sum w^{1,1}_{ij}X^1_{j}$, $\sum w^{2,1}_{ij}X^2_{j}$  respectively. Then after the non-linearity, we get $$\sum w^{1,1}_{ik}w^{1,1}_{il}X^1_{k}X^1_{l}$$
These are then concatenated across modalities, and passed through another multiplicative non linearity. Hence the features obtained after this layer are given by $$\sum w^{1,1}_{ik}w^{1,1}_{il}w^{2,1}_{i'm}w^{2,1}_{i'n}X^1_{k}X^1_{l}X^1_{m}X^1_{n}$$

Effectively we have a linear combination of degree 4 terms that are symmetric in modalities i.e. $\{ X^1_iX^1_jX^2_kX^2_m$, $X^2_iX^2_jX^2_kX^2_m$,  $X^1_iX^1_jX^1_kX^1_m \}$

However in the late fusion design the unimodal features are concatenated first, and then passed through the non-linearity. This produces after the first layer features of the form
$$ w^{1/2,1}_{ij} X^{1/2}_iX^{1/2}_j$$ 
where $w^{1/2}$ and $X^{1/2}$ mean that choice over modalities 1 and 2 can be applied to $w$ and $X$ respectively. More simply one gets all pairwise terms from both modalities, instead of pairwise terms from only individual modalities. The next non-linearity, produces further multiplicative terms and we get linear combination of all degree 4 terms i.e. $X^{1/2}_iX^{1/2}_jX^{1/2}_kX^{1/2}_m$

Note here that no choice of linear operators between the layers can produce non-symmetric cross modal terms in the late fusion;
and hence early fusion model has access to more features.
On the other hand the first non-linearity in the early fusion case scales quadratically in the number of modalities. Specifically if the input dimensionality of each modality is $D$, and the number of modalities is $n$, then in early fusion the first non-linearity produces $\Mycomb[nD]{2}$ feature outputs, whereas late fusion produces $n\Mycomb[D]{2}$. In the next non-linearity where a second multiplicative pairing occurs, early fusion needs ${}^{\Mycomb[nD]{2}}C_2$ parameters whereas late fusion needs ${}^{n\Mycomb[D]{2}}C_2$. As such one might need many more samples to learn an early fusion model compared to a late fusion approach.

The back-projections of Pro-fusion model however alleviates the lack of feature diversity in late fusion. The back-projections provides access to some cross-modal features before the fusion layer. This allows pro-fusion access to any given single asymmetric degree 4 term; however due to the limited dimensionality of the backprojected activations, not all combination of such degree 4 terms are accessible. For example if the backprojecting layer has size 2, then one can get only two pairs of independent asymmetric cross modal feature terms. Hence pro-fusion is more expressive pro-fusion but not as rich as early fusion.

\section{Training Backprojection layers}
For training of backprojecting layers, we first build an augmented model $\hat{\mathcal{F}}$ by first extending the unimodal feature generators in the the base model. Next we add the backprojecting networks $W_i$ for each modalities. We pass the fused output to the unimodal generators through the backprojecting networks. Finally we fix a number of iterations, and unroll the model by applying the augmented network in a loop. The entire process is now differentiable, and autograd can compute the gradients.

\begin{lstlisting}[basicstyle=\scriptsize,language=Python]
# base model is assumed given in the layers
class FusModel(nn.Module):
    def __init(unimodal_layers, fusion_layer, hp):
        self.unimodals = nn.ModuleList(*unimodal_layers)
        self.fusion_layer = fusion_layer
        self.num_modalities = len(unimodal_layers)
        
    def forward(inputs):
        uni_reps = [self.unimodals[_](inputs[_]) for _ in range(self.num_modalities)]
        fused_rep = self.fusion_layer(uni_reps)
        return fused_rep
        
class LateFusModel(nn.Module):
    def __init(unimodal_layers, fusion_layer, head, hp):
        self.base_model = FusModel(unimodal_layers, fusion_layer, head, hp)
        self.head = head
        self.num_modalities = len(unimodal_layers)
        
    def forward(inputs):
        fused_rep = self.base_model(inputs)
        return self.head(fused_rep)
    

class ProFusModel(nn.Module):
    def __init(num_modalities, unimodal_layers, fusion_layer, head, hp):
        self.back_projecting_layers = nn.ModuleList([self.build_back_layers(unimodal_layers[i], fusion_layer.output_dim,  hp) for i in num_modalities])
        self.augmented_readers = nn.ModuleList([self.extend_layer(unimodal_layers[i], hp) for i in  num_modalities])
        self.model = FusModel(self.augmented_readers, fusion_layer, hp)
        self.output_head = head
        self.hp = hp
        
    def forward(self, inputs):
        context_t = torch.zeros([inputs[0].shape[0], self.context_size])
        context_t = [self.back_projecting_layers(context_t) for _ in range(self.hp.num_modalities)]
        for _ in range(self.hp.num_unroll_steps):
            context_t = model_out = self.model(inputs, context_t)
            context_t = [self.back_projecting_layers(model_out) for _ in range(self.hp.num_modalities)]
    return self.output_head(model_out)

def train_epoch(model, optimizer, criterion, hp, train_loader):
    for i_batch, batch_data in enumerate(train_loader):
        modalities, tgt = batch_data
        optimizer.zero_grad()
        preds = model(modalities)
        loss = criterion(preds, tgt)
        loss += model.regularizer(hp)
        loss.backward()
        optimizer.step()
\end{lstlisting}

\newpage

\begin{figure}
    \centering
    
\includegraphics[width=0.4\textwidth]{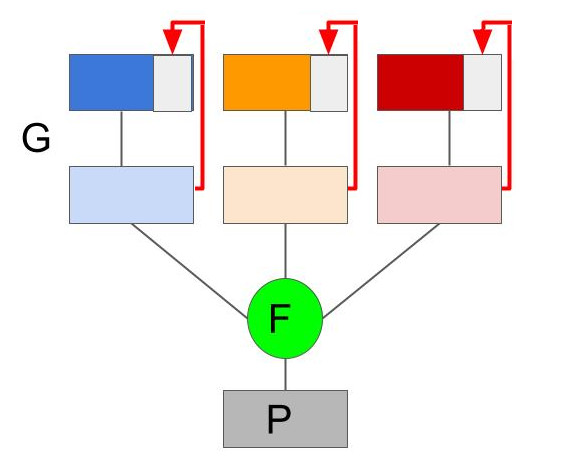}
\captionof{figure}{Example unimodal iterative network for ablation against ProFusion
\label{fig:apx:ablation_unroll}}
\end{figure}

\section{Exploring the Representations}
\label{apx:ablation2}


\begin{figure}[ht]
\centering
    \begin{subfigure}[b]{0.4\textwidth}
    \centering
    \includegraphics[width=\textwidth]{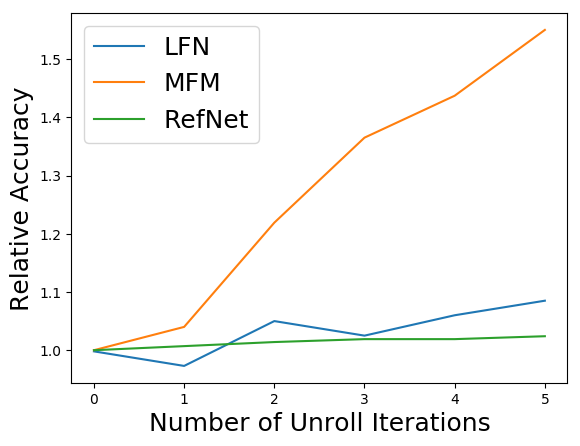}
    \caption{Accuracy from audio representation \label{fig:unimodal:audio}}
    \end{subfigure}
    ~
    \begin{subfigure}[b]{0.4\textwidth}
    \centering
    \includegraphics[width=\textwidth]{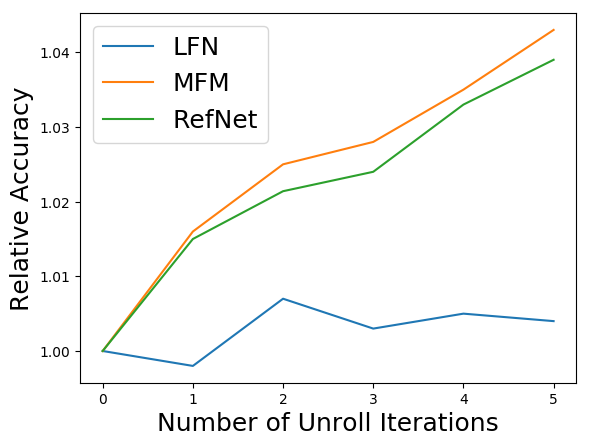}
    \caption{Accuracy from image representations  \label{fig:unimodal:img}}
    \end{subfigure}
    
    \caption{Behavior of classifiers trained on the unimodal embedding against number of unrolling iterations. We train a linear model on the input to the fusion layer and on them. The lines plot the normalized accuracy of the trained model.  We observe \emph{increased accuracy} with more unrolling.}
    \label{fig:unimodal_vs_K}
\end{figure}



Next, we analyze how the unimodal representations evolve over the unrolling steps. For this purpose, we consider the activations of unimodal networks $\hat{G}_j$ (equivalently, the inputs for the late fusion layer) as the unimodal representations. For these of experiments, we use LFN, MFM and Refnet models on AVMNIST. We train a linear classifier based on the unimodal representations from the training data and find its accuracy on the test data.

In Figure \ref{fig:unimodal_vs_K} we plot the relative test accuracy of both the audio and image features against the iteration number for all the models. We can see gains in all models after one step of unrolling. Since we know that individual modalities are quite incomplete on AVMNIST (< 60\% accuracy on individual modalities) and the accuracy of only image modality at the first step is close to 60\%. 
This suggests that the unimodal modalities are now integrating information from each other.  Along with the overall trend, this suggests that \emph{the model is incorporating more multimodal information in each unimodal representation}. 

\end{document}